\documentclass[10pt,twocolumn,letterpaper]{article}
\usepackage{times}

\usepackage{iccv}
\usepackage{times}
\usepackage{epsfig}
\usepackage{graphicx}
\usepackage{amsmath}
\usepackage{amssymb}
\usepackage{caption}
\usepackage{subcaption}
\usepackage{float}
\usepackage{algorithm}
\usepackage[noend]{algpseudocode}
\usepackage{bbm}

\usepackage[pagebackref=true,breaklinks=true,colorlinks,bookmarks=false]{hyperref}

\def\iccvPaperID{2500} 

\iccvfinalcopy 
\DeclareMathOperator*{\argmin}{argmin}
\begin{document}

\title{Geometric Affordances from a Single Example via the Interaction Tensor}

\author{Eduardo Ruiz, Walterio Mayol-Cuevas \\
Department of Computer Science\\
University of Bristol, UK\\
{\tt\small \{er13827,wmayol\}@bristol.ac.uk}
\and
}

\maketitle

\begin{abstract}
This paper develops and evaluates a new tensor field representation to express the {\it geometric} affordance of one object over another. We expand the well known bisector surface representation to one that is weight-driven and that retains the provenance of surface points with directional vectors. We also incorporate the notion of {\it affordance keypoints} which allow for faster decisions at a point of query and with a compact and straightforward descriptor. Using a single interaction example, we are able to generalize to previously-unseen scenarios; both synthetic and also real scenes captured with RGBD sensors. We show how our interaction tensor allows for significantly better performance over alternative formulations. Evaluations also include crowdsourcing comparisons that confirm the validity of our affordance proposals, which agree on average 84\% of the time with human judgments, and which is 20-40\% better than the baseline methods.
\end{abstract}
\let\thefootnote\relax\footnote{ER is supported by the Mexican Council for Science and Technology (CONACYT) scholarship scheme.}
\section{Introduction}
Perhaps the most fundamental question about Vision is {\it what is it for?} From the early propositions to computationally address this by D. Marr \cite{Marr1982}, the path has often been assumed to aim to, or at least require to, recover geometric information about the environment. This has directed much effort towards a particular spatial representation of the world, but at the same time, has moved effort away from the rationale that drove Vision to its high level of utility. 

\begin{figure}[ht]
        \centering
        \includegraphics[width=0.45\textwidth]{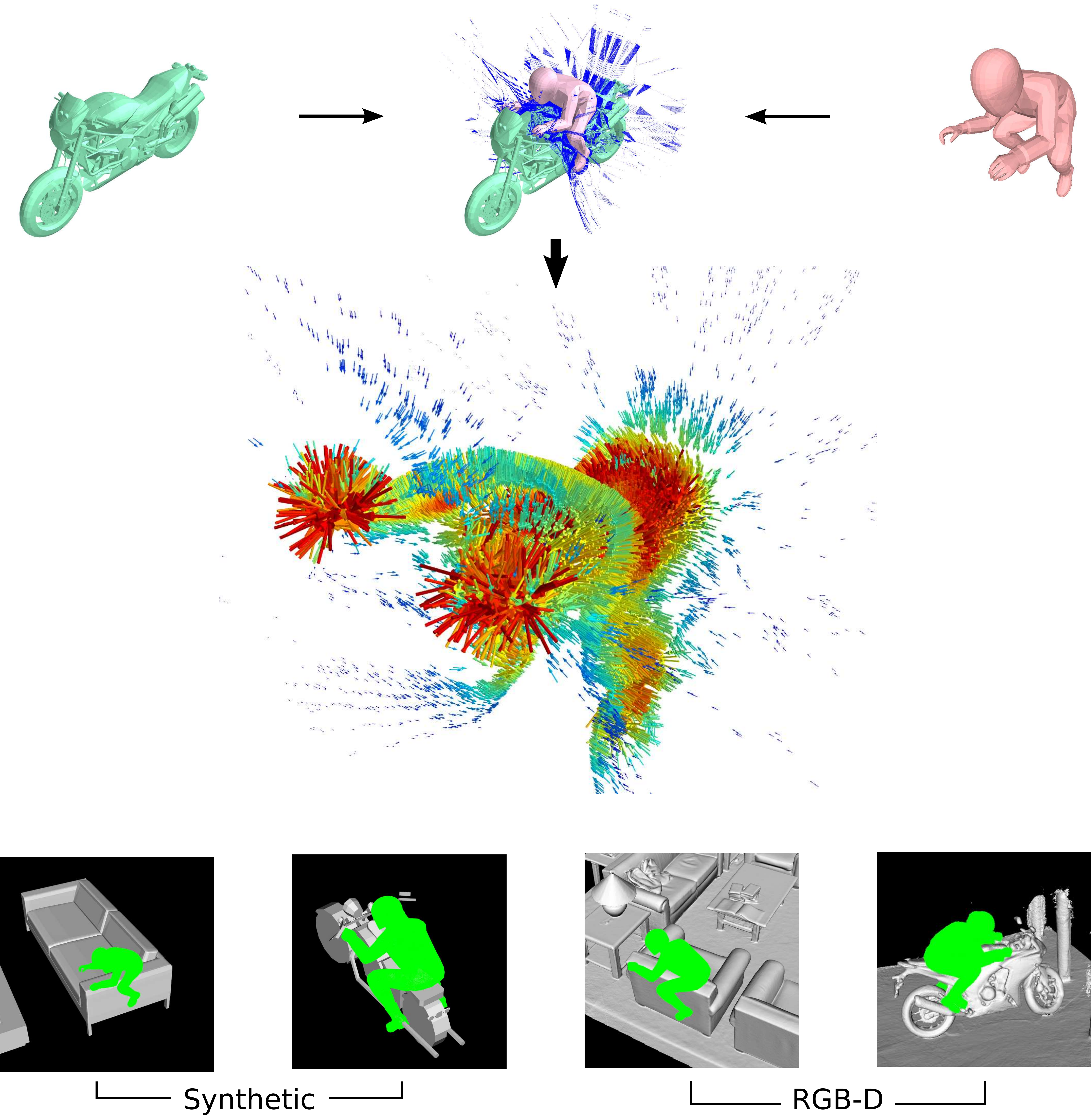}
      \caption{Our affordance tensor (centre) describes the interaction pair of "riding" (top row). This allows us to predict affordance locations on previously unseen scenes (bottom row), even under changes of geometry and from a single example. The geometric affordance we estimate agrees with judgments of mechanical turk markers, and is able to answer something like: {\it where would the kids pretend to ride a motorbike on the living room?}}
        \label{fig: general view}
        \vspace{-3mm}
\end{figure}

The view posed by J.J. Gibson however \cite{gibson1977}, calls for a visual perception that is there to help the perceiving agent to interact with the world. Specifically, through the coining of the term {\it affordance}, visual perception is described as a process to understand what can be done where. Such a representation of the world is immediately useful as by definition it is one that already takes into account what the agent is capable of.

Gibson also argued that affordances are "immediate" to perceive. This has often been misread as a call to ignore the relevance of the representation \cite{Warren2012}. But direct perception of affordances does not mean that no intermediate processing should take place, including perhaps requiring a 3D reconstruction; for instance, it has been shown that the dorsal stream in the visual cortex of the brain, computes edge detection, depth, 3D surface and axis representations \cite{Oztop2006}. Importantly, we argue that, the direct nature of affordance perception rather motivates methods that are able to immediately transfer what has been learned to other objects and places after a small number or even a single observation of the affordance.

Being able to determine affordances can have profound implications for visual systems. It can in principle liberate the computational approach to visual processing from the focus on objects and their arbitrary labels which have to be extensively learned. To learn an affordance is not to classify an object \cite{gibson1977}, since a cup is not only for drinking but also a paperweight, or even a tool to build sand castles.

While the concept of an affordance can appear elusive to express, an affordance is necessarily the result of the composition between the world and the agent. And one that is ultimately designed to be useful for the perceiving agent. Understanding and modeling this interaction between agent and the world is of central focus to our work and the one that we aim at.

We here concentrate on the subclass of affordances between rigid objects. Affordances such as where can I hang this?, place this, ride, fill, and similar. We do this by specifying a geometry-driven interaction tensor that aims to capture the way in which the affordance manifests between a pair of objects. 

Importantly, using only a single example, we then detect other viable places for such affordances in previously unseen scenes. We evaluate with both synthetic and real scenes.

Our approach is inspired by the well established concept of bisector surfaces (see \cite{Peternell2000} for an introduction), and their recent exploits for scene indexing \cite{Zhao2014}.
Here we extend these concepts by directly enhancing the bisector surface points with rational weighting, provenance vectors and the concept of affordance keypoints. All which results in a richer vector field or tensor. Our contributions in this paper can be outlined as follows:

\begin{itemize}
\item We extend the notion of the bisector surface to a weighted vector field ---an interaction tensor field.
\item Show how this tensor with direct, sparse sampling, allows for the determination of geometrically similar interactions even from a single example, and is better than existing formulations.
\item Introduce the notion of {\it affordance keypoints} which serve to more quickly judge the likelihood of an affordance at a query point.
\item Evaluate with both synthetic and real scenes from RGBD mapped areas.
\item We validate results with crowdsourced judgments.
\end{itemize}

\begin{figure}[ht]
        \centering
        \includegraphics[width=0.45\textwidth]{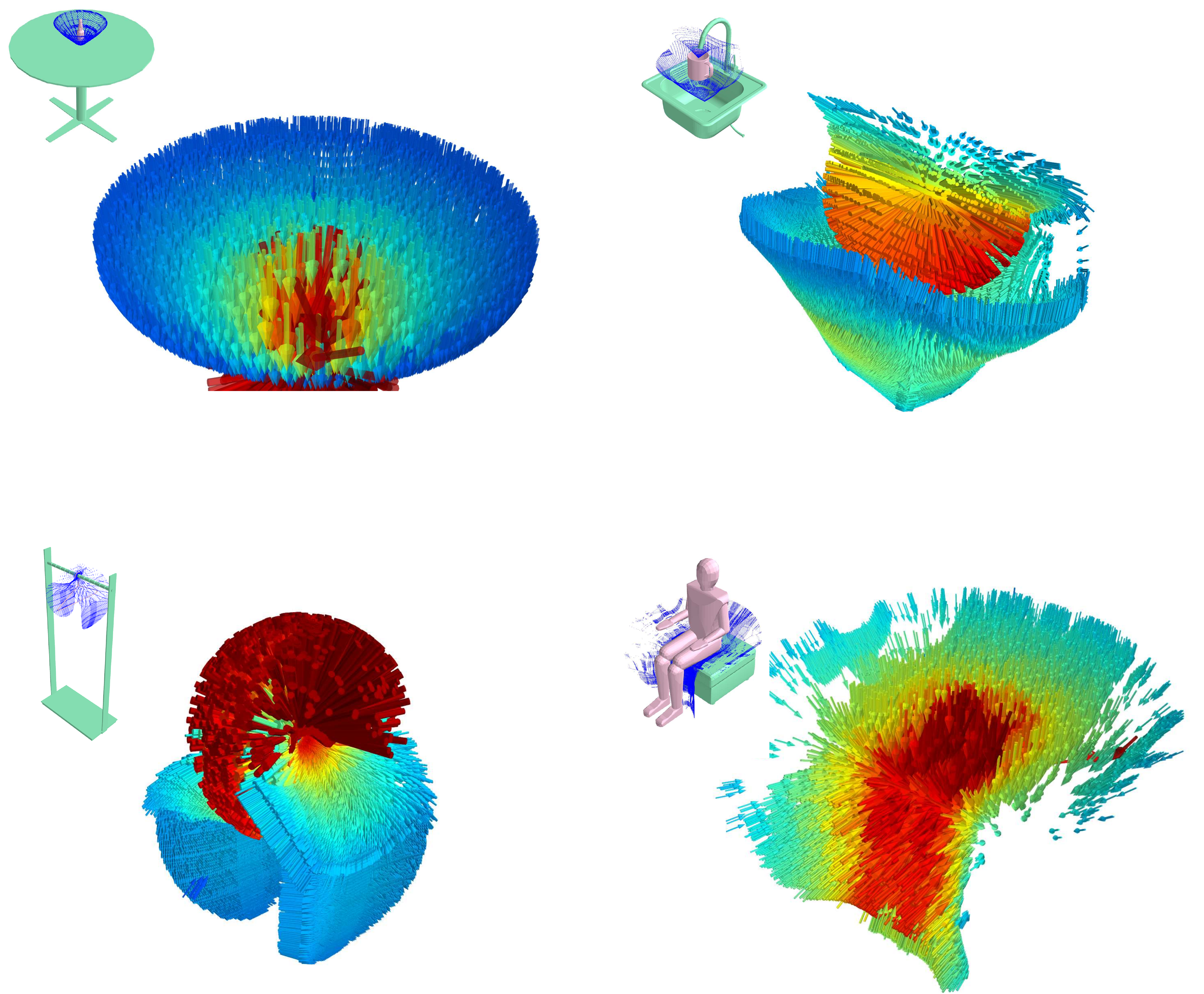}
      \caption{Interaction tensor examples of 4 affordances. Starting from the top-left in clockwise direction: \textit{placing} a bottle, \textit{filling} a mug, \textit{sitting} and \textit{hanging} a coat hanger.}
        \label{fig: surface_examples}
        \vspace{-3mm}
\end{figure}

\section{Related work}
Affordance detection has been studied in recent years in computer vision  and robotics. Briefly speaking, affordance knowledge has been incorporated in learning systems that use data from demonstrations of interaction, robot self exploration and static labeled imagery. In terms of the applications, the approaches include semantic scene understanding, grasp learning, gesture recognition, object segmentation and planning in goal-directed tasks.

An important body of research comes from the developmental robotics field \cite{Min2016}. The core of these approaches is the representation and learning of actions and predicting the consequences of these
over a set of objects. These approaches use visual features describing shape, color, size and relative distances to capture object properties and effects. Using robot self-exploration and human demonstrations the systems benefit from single-object affordances to execute more complex interactions and execute a plan (task planning). For instance, \cite{Mar2015} shows a robot learning in a self-supervised manner to use a tool by observing the effects of its actions on other objects.

Another line of research that has benefited from affordance learning is Human-Robot Interaction \cite{saponaro2013,pandey2013,pieropan2013,koppula2014,srikantha2014,chan2014,jiang2014}. In these studies 
the main goal is to to perform action recognition in a robot observing humans, usually to predict or anticipate human activities, and in this way assist humans better while they perform everyday tasks.

Work has also been done using static imagery, where the affordance or interaction is provided as a label rather than demonstrated. \cite{desai2013,srikantha2014,Zhu2014,Chao2015}
based their work on labeled 2D imagery to predict functional regions or attributes on every day objects.

A body of research closer to our approach is the one exploiting 3D information to learn and predict affordances of objects in the environment. In \cite{aldoma2012}, the concept of 0-order affordance is introduced to refer \textit{hidden} affordances 
that can be found on an object but not in its current pose. Amongst the affordances studied are rollable, containment, liquid-containment, unstable, stackable-onto and sittable. In \cite{hinkle2013} 
a physics-based simulation on CAD models of objects is used to learn three functional classes: drinking vessel, table and sittable. Using geometric features on RGB-D data \cite{kim2014} presents a segmentation 
algorithm that learns and predicts affordances such as pushable, liftable and graspable on indoor scenes. In \cite{Myers15,Nguyen2016} RGB-D images are used to learn and predict functional regions such as grasp, contain,
support and cut on objects placed on a table-top. Using RGB-D images of indoor scenes \cite{Roy2016} perform segmentation for human actions such as walkable, sittable, lyable. Similarly, affordances are studied in \cite{Piyathilak2015,Gupta2011,jiang2014} to map locations suitable for sitting, or laying down; particularly in these cases using human skeleton \textit{hallucinated} on the different indoor scenes.
Crucially, these previous methods are heavy in terms of requiring multiple learning examples, impose a particular parameterization such as detection of planes or shapes and or are highly specific to an object \eg humanoid shapes. Our approach aims to address various of these limitations, namely relying on pre-parameterization of scene or objects and relying in numerous examples.

\begin{figure*}[ht]
        \centering
        \includegraphics[width=0.98\textwidth]{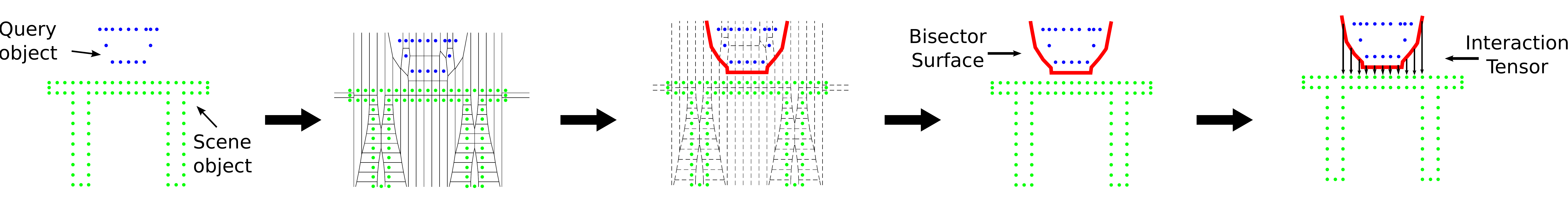}
	\caption{The interaction tensor is computed from the bisector surface. First, objects are placed simulating the interaction. The Voronoi diagram is calculated amongst all the data points. Only ridges splitting points from different objects are taken into account. These points comprise the bisector surface (red), which is used to compute the interaction tensor for \textit{placing} a bowl on a table.}
\label{fig: explanation}
\vspace{-3mm}
\end{figure*}

\section{Our approach}
As noted before, geometrical properties are very important in the modeling or representation of affordances. And this is validated by success of some of the related methods.


An interesting additional related work is \cite{Zhao2014}, where an algorithm for 3D scene indexing is developed to capture hierarchical relationships among objects using Betti numbers. It proposes Interaction Bisector Surface curvature descriptors that are learned from multiple examples. There, it is shown the discriminative power of the Bisector Surface (BS) to characterize the relationships between sets of objects. The bisector surface $B$ of two objects $O_1,O_2$ is the locus of points equidistant to the objects' surface. The bisector surface is an approximation of the Voronoi diagram for objects in a scene. 


We extend the robustness of the BS by preserving information regarding the expected locations or areas in the 3D space that enable the interaction. This is what we call the Interaction Tensor (iT).

Furthermore, we are able to identify areas of high importance for each affordance based on the geometric and spatial relationships between the interacting objects. 
Briefly speaking, our method consists on computing the iT descriptor between a pair of objects of whose affordance we are investigating. Examples of the iT between pairs of objects are shown in Fig. \ref{fig: surface_examples}.


As first step, the objects are placed simulating the interaction that they would have on real circumstances (affordance example); we then compute the BS produced by 
these two objects and preserve the \textbf{provenance vectors}. That is, the vector(s) that contributed to the computation of a given point in the BS. But note that the provenance vectors should not be confused with surface normal vectors on the BS, since the latter do not provide information regarding from where the BS points actually come from. This process generates the Interaction Tensor for the affordance simulated by the two interacting objects. At test time, we are able to predict affordance location candidates by approximating the iT on a previously unseen input scene. The method allows us to use a model of say a humanoid \textit{skeleton} and predict \textit{human affordances} such as sitting; similarly to  \cite{Gupta2011,jiang2014,Piyathilak2015,Roy2016}. But importantly, it also allows us to build these tensors more generally for any other pair of objects.

Formally, given a bisector surface $B$ formed by points $b$ and an object $O$ in the scene formed by points $o_i,...,o_n$, the tensor field characterizing the interaction is defined as
\begin{equation}
\mathbf{iT}(B) = P\hat{i} + Q\hat{j} +R\hat{k}
\end{equation}
where 
\[P = \hat{G}(B)_{\hat{i}}-B_{\hat{i}}
\]
\[Q = \hat{G}(B)_{\hat{j}}-B_{\hat{j}}
\]
\[R = \hat{G}(B)_{\hat{k}}-B_{\hat{k}}
\]
with
\[ \hat{G}(B)= \argmin_i \lVert o_i- B\rVert_2
\]

Additionally, we take into account the measurement of how important every location is in the iT. This is expressed as a weight as we will discuss later.



\subsection{Computing the Interaction Tensor}
The bisector surface between a pair of objects is computed similarly to \cite{Zhao2014}. Using 3D or CAD models of the interacting objects, the first step is to create dense point clouds by uniformly sampling points on the surfaces of the models. Then, the Voronoi diagram is computed for all these points; which produces a simplicial complex where polygon ridges are equidistant to the points that produced them. The Bisector Surface is comprised of ridges originated by points from different objects.
In our experiments, we refer to the two interacting objects as \textbf{query-object} and \textbf{scene-object} (or scene) respectively.
The query-object is the one with a known affordance; a mug which affords \textit{filling}, is an example of an affordance query in our setup. A scene-object is the second part of the interaction; this could be a second object or part of a scene or furniture that allows the affordance to take place. 
Using the same mug filling example, a faucet or tap and sink would act as scene-object. Fig. \ref{fig: explanation} illustrates how the interaction tensor is computed from the bisector surface between two sets of points. Specifically, it shows the iT for \textit{placing} a bowl on a table in a simplified 2D scenario. Recall we are concerned in this paper only in the geometric component of the affordance, that is, in things and places that {\it appear} to afford the task.

In principle, the BS and iT extend towards infinity; in practice, we trim these to fit a sphere of radius equal to the diagonal of the query-object bounding box.

\begin{figure}[ht]
\begin{center}   	
\includegraphics[width=0.45\textwidth]{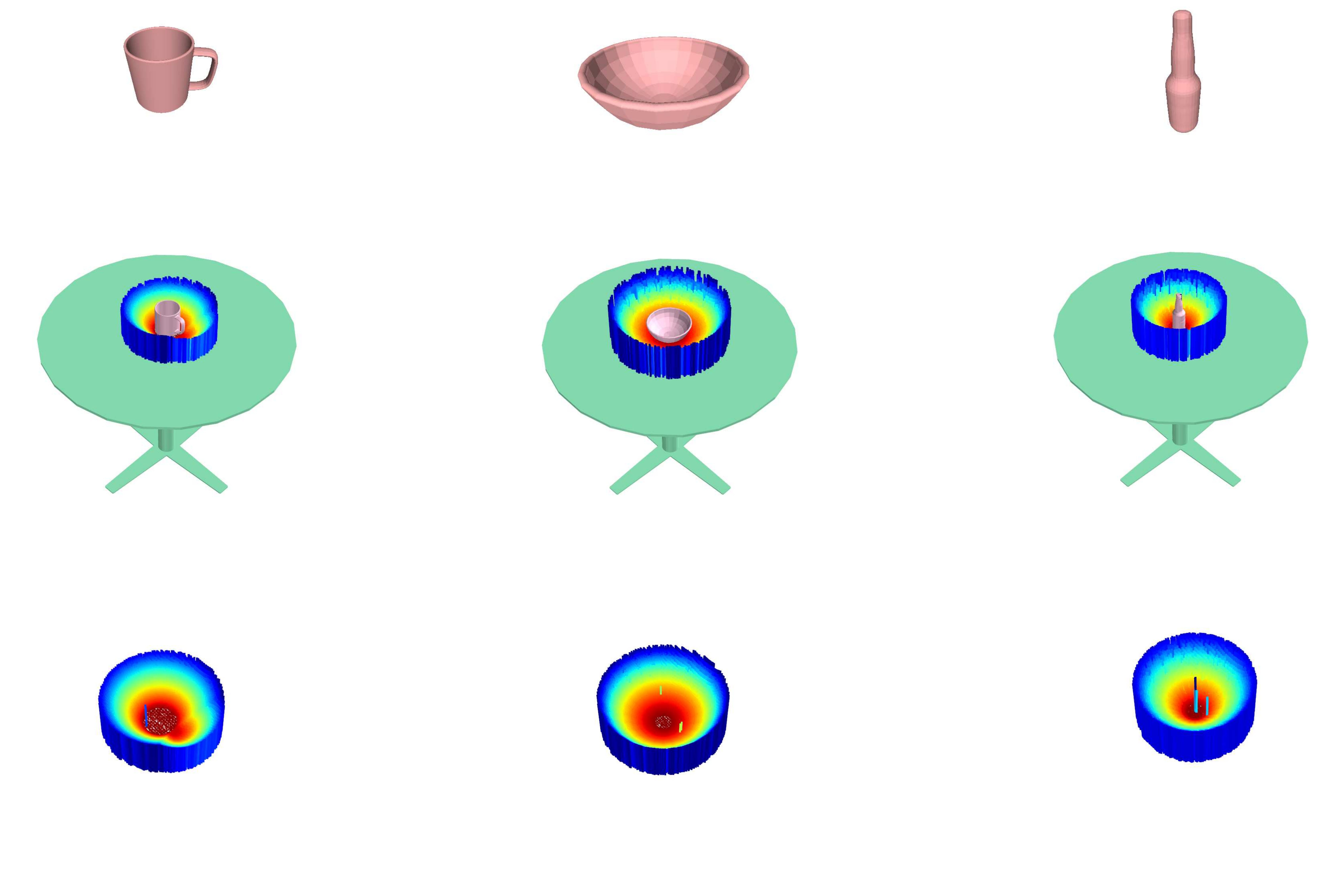}
	\caption{Examples of interaction tensor for the same affordance (\textit{placing}) using different query-objects: mug, bowl and bottle. The interaction tensors' similarity makes them robust to changes in geometry of the query-object.}
	\label{fig: surface_examples_placing}
\end{center}
\vspace{-6mm}
\end{figure}

\begin{figure}[hbt]
\begin{center}
\includegraphics[width=0.45\textwidth]{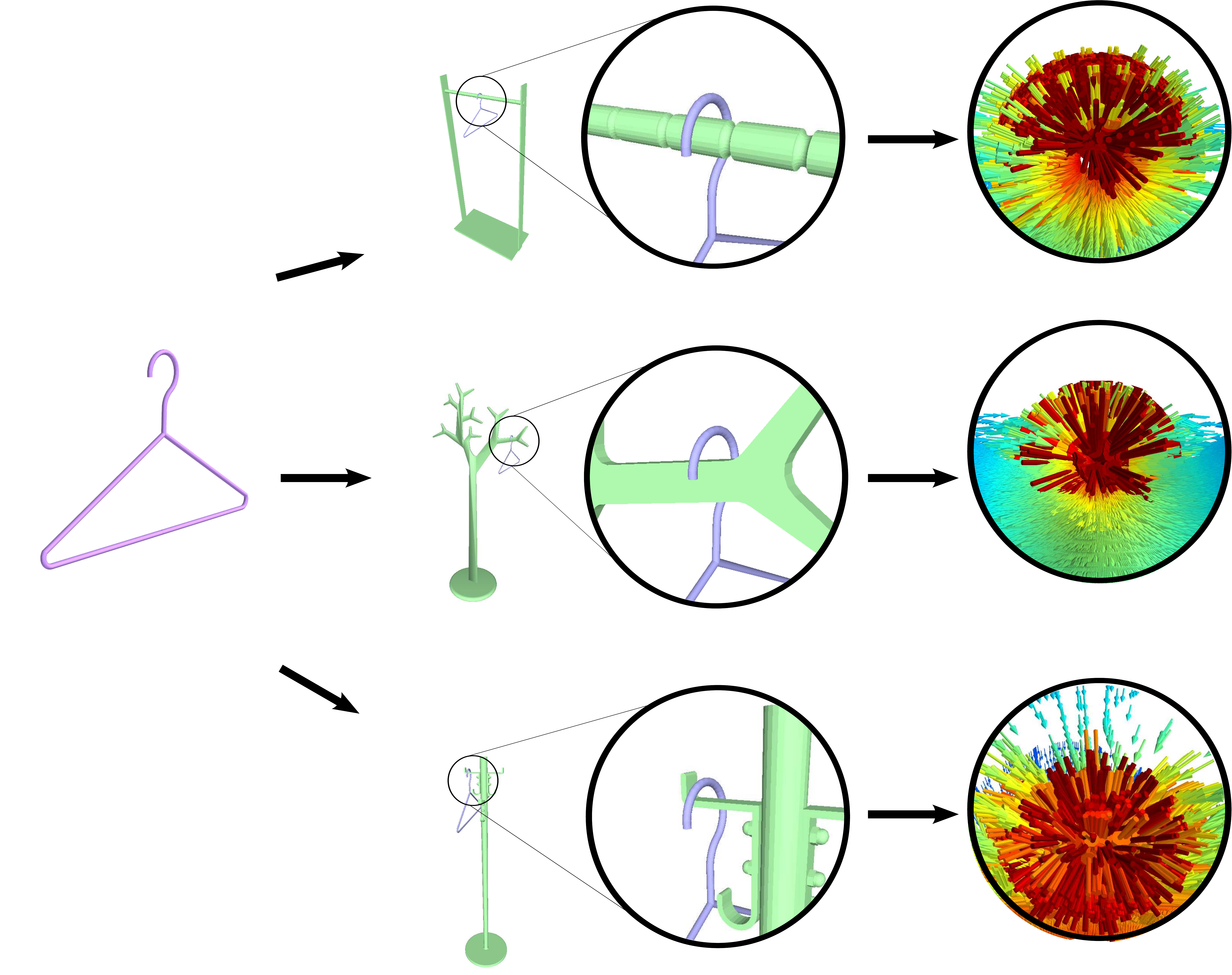}
 	\caption{Interaction tensor for \textit{hanging} a coat hanger on racks with different geometries. Although changes occur in specific locations of the tensor, the key features of the interaction are preserved. }
	\label{fig: surface_examples_hanging}
\end{center}
\vspace{-6mm}
\end{figure}

\begin{figure}[hbt]
\begin{center}
\includegraphics[width=0.47\textwidth]{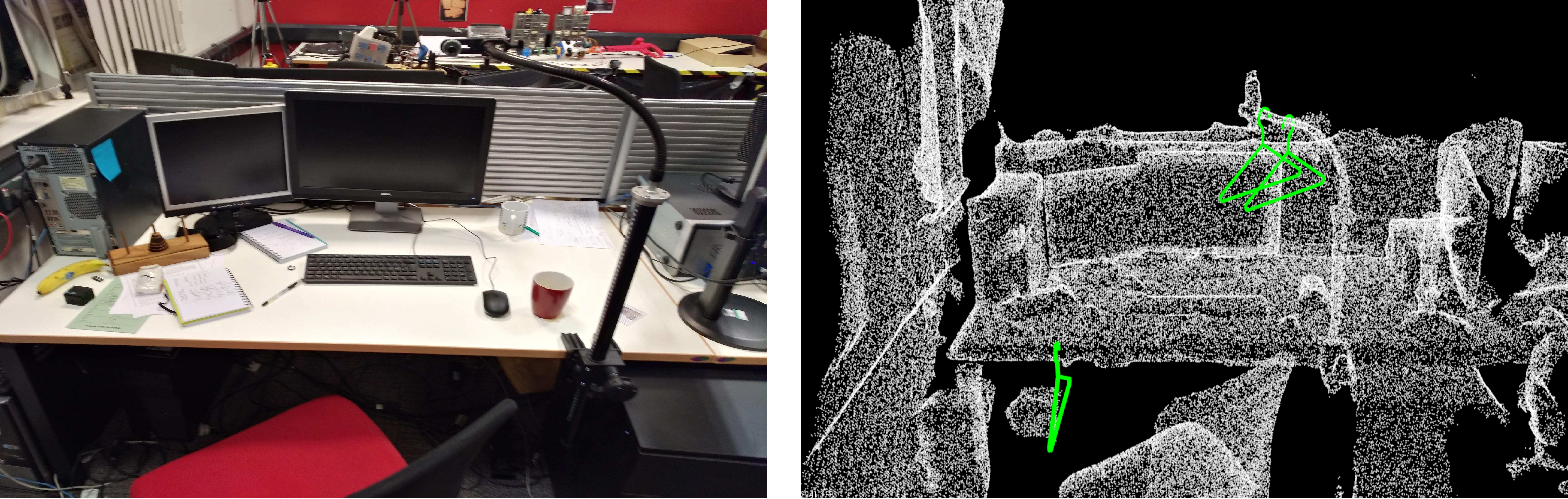}
 	\caption{Affordance prediction examples for \textit{hanging} a coat hanger in a real office-desk scene captured with an RGBD sensor.}
	\label{fig: real hanging}
\end{center}
\vspace{-5mm}
\end{figure}

The interaction tensor inherits from the bisector surface the discriminative power in characterizing the relationships between sets of objects. It preserves key geometrical features while being robust to changes in the geometry of the interacting objects. Fig.\ref{fig: surface_examples_placing} shows examples of the interaction tensor for the same affordance: \textit{placing} query-objects with changing geometries on a flat surface (table). Similarly, figure \ref{fig: surface_examples_hanging} shows interaction tensor examples generated using the same query-object (coat hanger) and scene-objects (coat racks) with varying geometries. In Fig.\ref{fig: real hanging}, the same single example affordance tensor learned from the synthetic scene is used on a real RGBD scene where meaningful placements are proposed.
These figures demonstrate that despite geometrical changes in the interacting objects the iT retains the overall shape or geometrical features characterizing the interaction. 

One single example interaction tensor is computed for every affordance considered in our research: \textit{placing}, \textit{hanging}, \textit{filling}, \textit{sitting} and \textit{riding}. Fig. \ref{fig: general view} shows the training example for \textit{riding}. Fig. \ref{fig: surface_examples} shows the interaction examples for the other 4 affordances.

\subsection{Weighted Interaction Tensor}\label{weights}

Every point in the BS is defined by a set \textit{provenance vectors}, we use this information to assign a weight to every location on the interaction tensor, $W=\{w_b , \forall b \in B\}$. 
We assume that scene-object's point clouds are dense enough, this allows us to take simply one of such vectors without losing generality. The weight related to a point in the interaction tensor is computed from the magnitude of its corresponding \textit{provenance vector}. This weight or distance, represents how relevant every point is for the interaction taking place between the objects. Fig. \ref{fig: general view} shows the weights for \textit{riding} a motorcycle. Fig. \ref{fig: surface_examples_placing} and \ref{fig: surface_examples_hanging} depict the weights for \textit{placing} and \textit{hanging}  affordances as the color of every vector in the interaction tensor. High weights are colored in red while lower weight locations are rendered in blue.

\begin{figure}[ht]
\begin{center}
    	\includegraphics[width=0.45\textwidth]{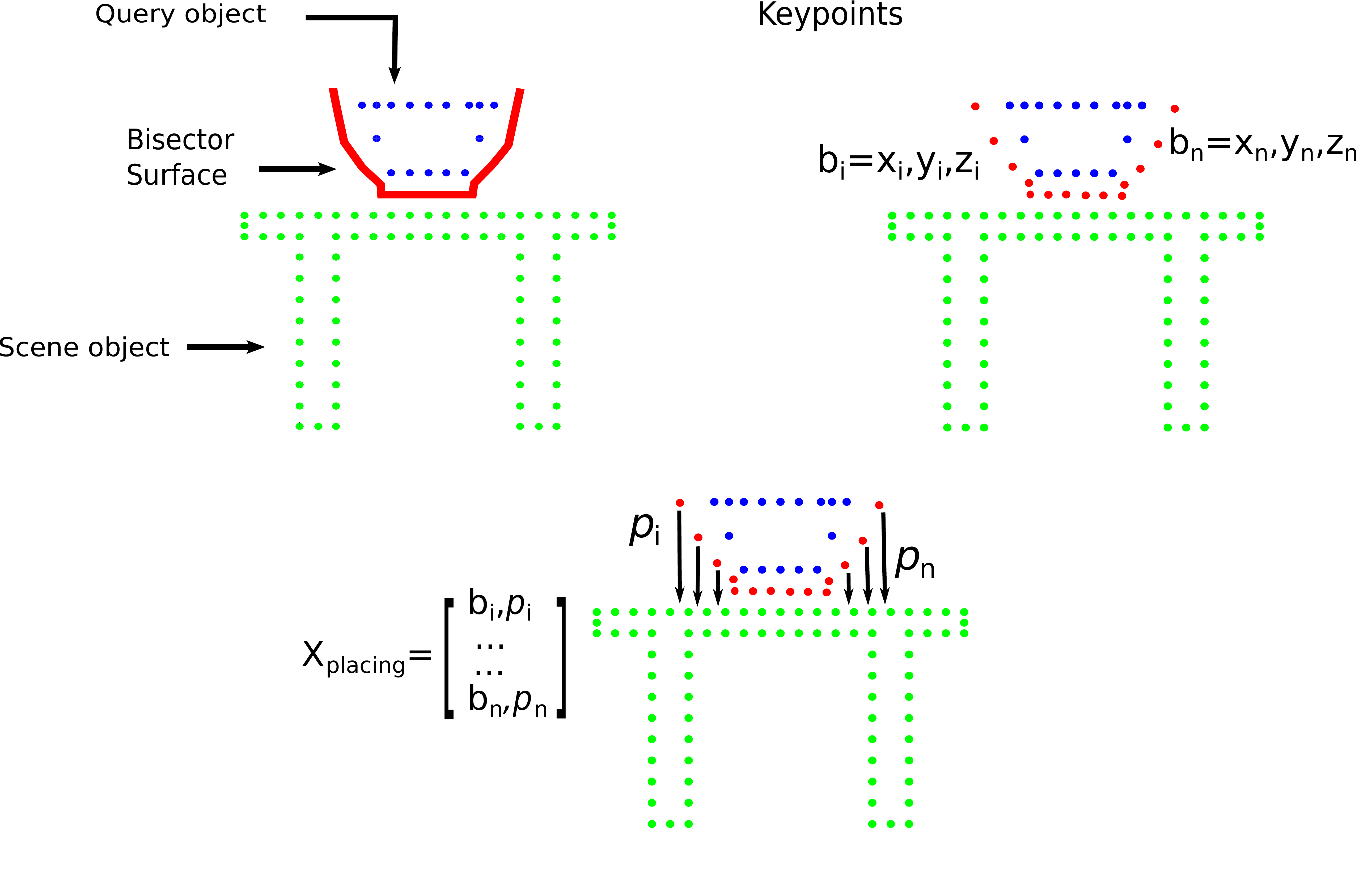}
	\caption{Affordance descriptor for \textit{placing} a bowl on a table in a 2D scenario. A set of points is sampled from the bisector surface. An affordance keypoint is obtained by computing the interaction tensor over these sampled points. These keypoints lead to the interaction tensor descriptor $\mathit{X}$.}
	\label{fig: descriptor}
\end{center}
\vspace{-3mm}
\end{figure}

The iT is a high dimensional and rich representation for object interactions, employing it directly as descriptor for affordance prediction would require costly computational resources. In order to reduce computational costs and improve the generalization capabilities of the descriptor, we reduce dimensionality by drawing N samples from $\mathbf{iT}$. This subset comprises what we call \textit{affordance keypoints} $\mathit{X}=\{ \mathit{X}_1, \mathit{X}_2,...,\mathit{X}_n\}$ where $\mathit{X}_i = \langle b_i, \mathbf{iT}(b_i)\rangle$. This lower-dimensional descriptor is formed by a set of points on the bisector surface and \textbf{provenance vectors}. In other words, each \textit{keypoint} is formed by a 6-dimensional feature vector which consists of the $x,y,z$ coordinates of the data point $b_i$ on the bisector surface, and vector $\vec{p}$ to its nearest neighbor in the scene-object $\mathbf{iT}(b_i)$. The scalar $\rvert \vec{p} \lvert$ encodes the importance (weight) of a keypoint in the interaction between objects; since in principle the shorter the distance between objects, the more significant the interaction is between them. Every \textit{provenance vector} $\vec{p}$ also suggests key locations in the scene that allow the interaction to take place. 
Fig. \ref{fig: descriptor} depicts graphically the method to compute \textit{affordance keypoints} forming the descriptor for \textit{placing} a bowl on a table in a 2D case.

Each affordance descriptor $\mathit{X}_{aff}$ has Nx6 dimensions, where N is the number of keypoints sampled from the iT (N=512 in our experiments). Two sampling
methods of the iT were tested: 1) uniform sampling and 2) weight-driven sampling. 
\textbf{Weight-driven} sampling uses weights $W$ from the tensor to form a probability distribution (\ref{eq:probs})
\begin{equation}\label{eq:probs}
 P(x_i \mid w_1,w_2,...,w_{n})=\frac{w_i}{\sum_{k=1}^{n}w_k}
\end{equation}
\begin{equation}\label{eq:weights}
w_i=1-\frac{\lvert \vec{v_i} \rvert -w_{min}}{w_{max}-w_{min}}
\end{equation}

Probabilities are inversely proportional to weights after normalization and weights are given by $\lvert \mathbf{iT} \rvert$. Equation \ref{eq:weights} ensures that the probabilities are in the range [0,1].
The idea behind this sampling method is to have a more meaningful representation (higher keypoint density ) in locations that are highly relevant for the interaction. These typically are locations where objects come closer together or touch, for instance in Fig. \ref{fig: general view} the saddle, handlebar grips and footrest of the motorcycle.

\subsection{Affordance query}

We are interested in predicting affordances or interaction possibilities on an input scene. Given a query-object and an affordance of interest, we predict good locations or possible places in the scene where the interaction could take place. Examples of such testing scenario are: ``where can I place a bottle?'',``where can I hang a handbag? or ``where can I fill a mug?'' 

Using these type of questions we perform a search over the input scene. Whereas this could be seen as an exhaustive process, we are able to prune the search by further characterizing each affordance with the expected orientation of the normal vector $N_{aff}$ at the surface or point in the scene-object enabling the interaction.  For instance, the \textit{filling} affordance requires normal vectors pointing downwards in the faucet. Interactions such as \textit{hanging} or \textit{sitting} require normal vectors pointing upwards, which in principle come from a surface supporting the query-object.

In order to make affordance location predictions we follow Algorithm \ref{alg:euclid}. First, points $p_i$ are randomly sampled all over the input scene (30\% of the total scene point in our experiments). Then, the normal vector $n_i$ is computed at that sample point; if this normal vector is similar to the expected $N_{aff}$ we extract a voxel centered at $p_i$ with a radius equal to the diagonal of the query-object bounding box $d_o$. 
From the \textit{training} example we have an approximation of the pose of $X_{aff}$ relative to the scene-object. The transformation corresponding to that pose configuration is applied to $X_{aff}$ to \textit{align} it as it would be expected if the affordance could take place at $p_i$. Using points within the current voxel as scene-object $O$, a nearest-neighbor search is performed for every point $b_i$ in $\mathit{X}_{aff}$ 
. With this information, vectors $\vec{v_t}=\overline{b_io_i}$ are computed at test time (\ie online), these are an approximation of \textit{provenance vectors} found in the iT example. Test vectors and example \textit{provenance vectors} are compared to obtain a score $s_i$. Using $n_i$ as rotation axis, scores are computed at different orientations $\theta$ 
(8 orientations evenly distributed in $[0,2\pi)$ in our experiments). Using an empirically tuned threshold $S_{aff}$ we are able to detect good matches (\ie affordance predictions) with the most likely orientation of the query-object.

\begin{algorithm}
\caption{Affordance query}\label{alg:euclid}
\begin{algorithmic}[1]
\ForAll{sample points $p_i$ in scene}
  \State Compute normal vector $n_i$
  \If{$n_i \approx N_{aff}$}
  \State Extract voxel of radius $d_o$
  \ForAll{orientations $\theta$}
    \State Compute score $s_i$ at $\theta_i$ using (\ref{eq:score})
    \If{ $s_i\geq S_{aff}$ }
      \State Predict ($p_i$,$\theta_i$) as good location
    \EndIf
  \EndFor
  \EndIf
\EndFor
\end{algorithmic}
\end{algorithm}

The function to compute the \textit{alignment} quality (\ie score) at a particular location at test time is shown below 
\begin{equation}\label{eq:score}
s_i=\sum_{i=1}^{N} \Theta(\vec{p},\vec{v_t})+\Delta(\vec{p},\vec{v_t})
\end{equation}
where 
\[\Theta(\vec{p},\vec{v_t}) =
 \begin{cases}
  1 & \quad \text{if } \angle (\vec{p} \vec{v_t}) \leq W\theta_{max} \\
  0 & \quad \text{otherwise}\\
 \end{cases}\]
 and
 \[\Delta(\vec{p},\vec{v_t})=
 \begin{cases}
  1 & \quad \text{if } |\vec{p}|-|\vec{v_t}| \leq W\frac{|\vec{p}|}{c} \\
  0 & \quad \text{otherwise}\\
 \end{cases}\]
Where $W=w_{max}/w_i$, $\theta_{max}$ is the maximum angle difference allowed, $c$ controls the maximum difference in magnitude between \textit{provenance vectors} and test vector as a proportion of the expected distance $|\vec{p}|$.  We empirically set $c=5$ and $\theta_{max}=\frac{pi}{18}$; meaning that on more significant keypoints (highest weight) the scoring criterion in more strict and differences should not be greater than 20\% of the expected values (iT example). In a first step the angle difference is computed, only if the difference between angles ($\Theta$) is small enough the magnitudes ($\Delta$) are compared.
 
\section{Experimental Results}
\label{experiments}

\subsection{Synthetic data}
\begin{figure}[ht]
        \centering
	\includegraphics[width=0.45\textwidth]{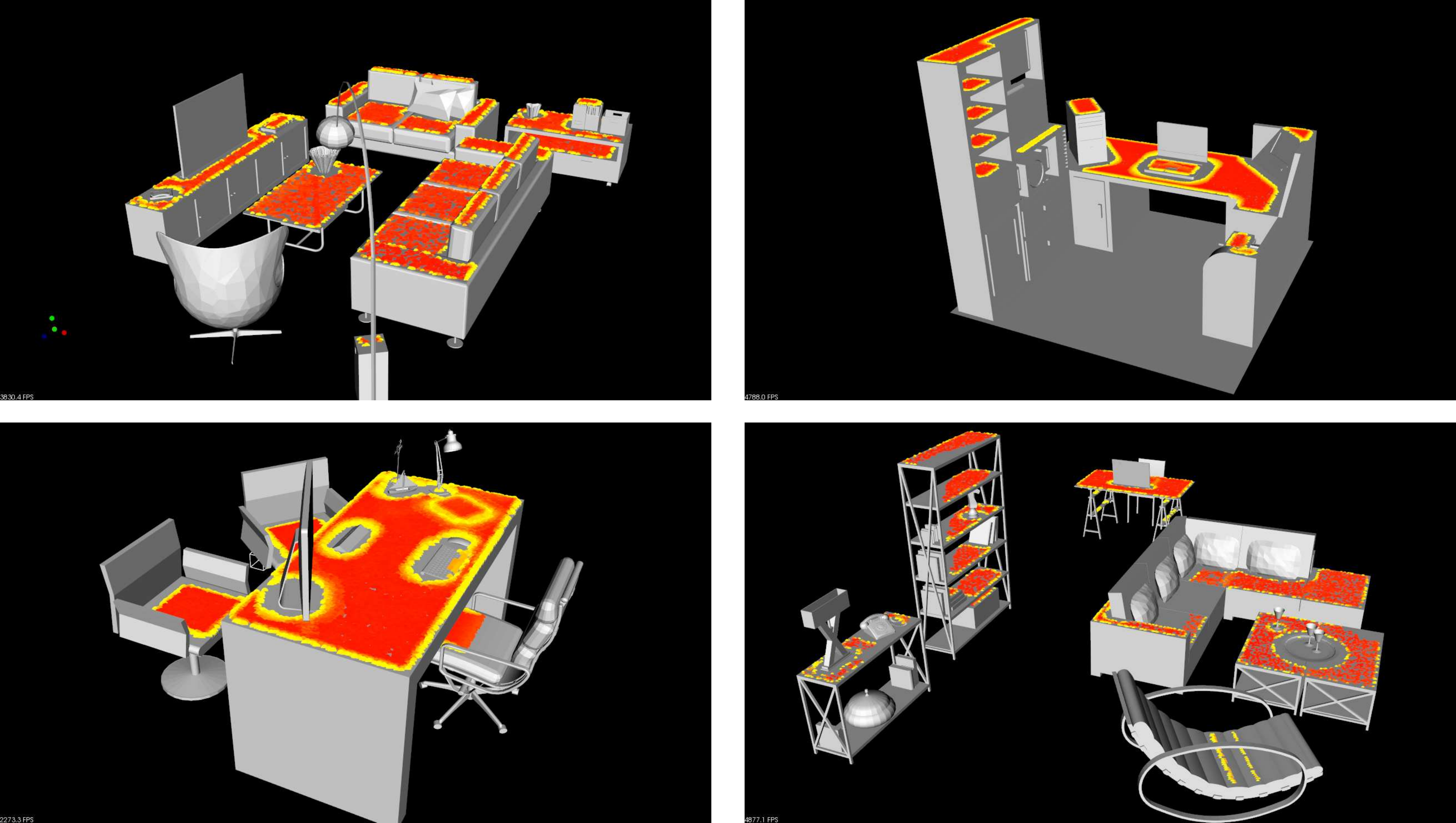}
      \caption{Example of synthetic scenes in our affordance prediction experiments. They show the prediction heat-map produced by our algorithm. The complete dataset is comprised by 5 kitchens, 5 living rooms and 5 office spaces.}
        \label{fig: scenes}
\end{figure}

For our experiments, we considered a total of fifteen synthetic scenes: 5 living rooms, 5 kitchens and 5 offices; and 8 affordance-object pairs \textit{filling-mug, filling-cup, placing-bottle, 
placing-bowl, hanging-hanger, hanging-handbag, sitting-human, riding-human}. Fig. \ref{fig: scenes} show examples of the scenes that we have considered and the output affordance 
heat-map obtained using our algorithm. All the CAD models (objects and scenes) were publicly available from the Trimble 3D warehouse \footnote{https://3dwarehouse.sketchup.com/}.

Due to space limitations we only show results from a subset of our scene dataset. But more data is available upon request and in the supplementary material.

\subsection{Evaluation}

\begin{figure*}[ht]
        \centering
	\includegraphics[width=\textwidth]{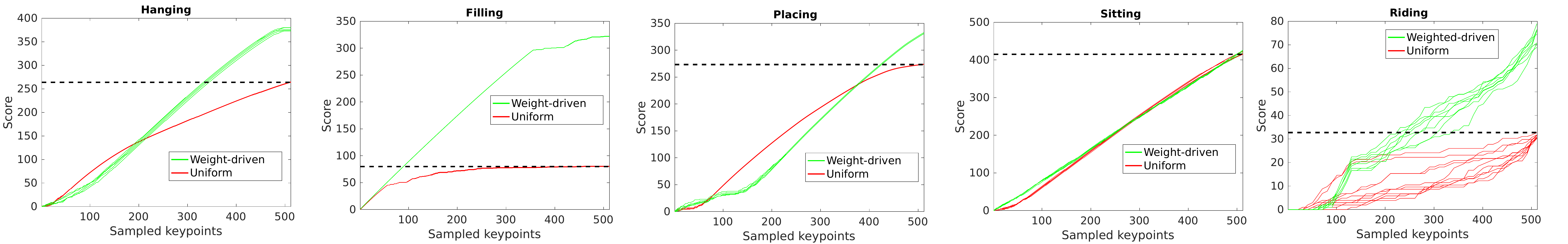}
      \caption{Plots show performance of two keypoint sampling methods. For most affordances weight-driven sampling achieves the prediction score threshold faster than uniform sampling (less comparisons made at test time). For some affordances the difference can be subtle, whereas in some others such as \textit{filling} affordance, the difference goes to 80\%.}
        \label{fig: sampling comparison plots}
        \vspace{-3mm}
\end{figure*}

\subsubsection{Sampling methods}
In a first set of experiments we tested our sampling methods: 1) Weight-driven and 2) Uniform sampling. Fig. \ref{fig: sampling comparison plots} shows the performance achieved for the 5 different affordances considered in our research. The plot shows the scores of the top 20\%  affordance predictions made by both methods. When weight-driven sampling is used to compute scores, the algorithm reaches the prediction score threshold faster than uniform sampling in four out of 5 affordances. This is due to the fact that by making sure that higher weight keypoints are present the algorithm is confident enough to make a prediction. High-weight keypoints are compared first by the algorithm, hence the score threshold is achieved performing less computations. On the other hand, uniform sampling achieves similar performances using on average 40\% more comparisons (\ie 200 keypoints more); which in most cases means all the keypoints in the descriptor. Qualitatively, both methods provide similar results, Fig \ref{fig: ibs comparison}a and \ref{fig: ibs comparison}b show this comparison for \textit{hanging} a coat-hanger.

\subsubsection{Interaction Tensor vs baselines}
\begin{figure}[ht]
        \centering
	\includegraphics[width=0.45\textwidth]{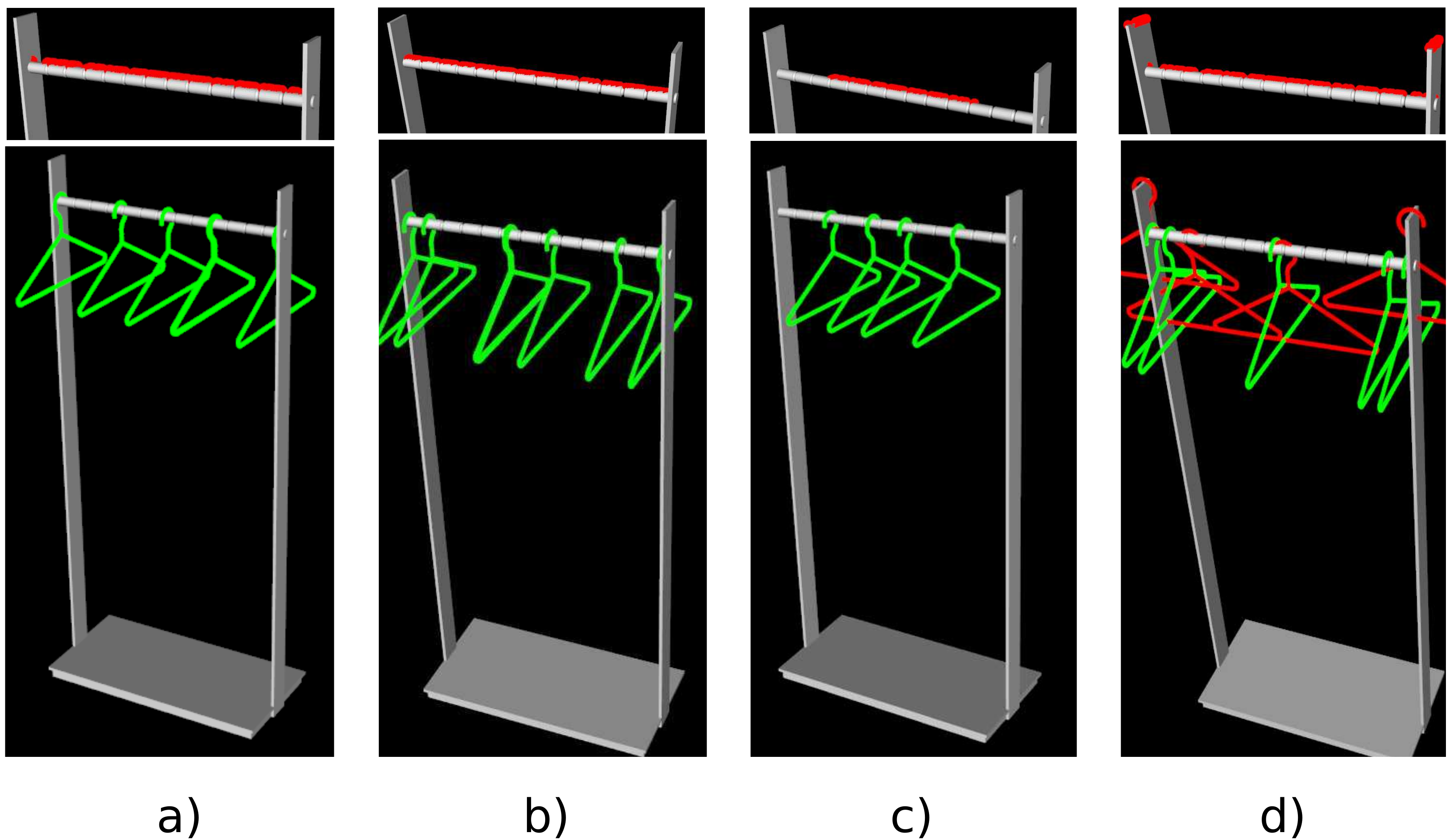}
    \caption{The iT descriptor allows more flexibility in the prediction of affordance location candidates using uniform sampling (a) and weight-driven sampling (b). The BS (c) predicts affordance location closely similar to the \textit{training} example (center of the hanging rack). In order to  achieve similar performance with BS the similarity threshold has to be relaxed (d), but at the expense of increasing the number of false positives (red coat hangers).}
        \label{fig: ibs comparison}
        \vspace{-3mm}
\end{figure}

First we compare the performance of our approach against using the BS as descriptor. Using ICP (\cite{Rusu2011} implementation), a score is computed between the BS from the interaction example and the one computed at test time. In addition to being slower or more computationally intensive, the BS descriptor is much more strict by trying to find interaction opportunities only closely similar to the \textit{training} example. One first advantage of our approach is that, by considering a weighted vector field, we have a more relaxed matching criterion in parts of the interaction that are not critical to the affordance; this allows us to detect affordance locations in spite of variations in the scene geometry while remaining robust against false positives. In order to achieve a performance similar to the iT descriptor, it is necessary to relax the matching threshold for BS comparisons; however, this increases the number of false positives. Fig. \ref{fig: ibs comparison}d shows an example of such circumstances for \textit{hanging} a coat-hanger on a rack.

\begin{figure}[ht]
    \centering
    \begin{subfigure}[b]{0.46\textwidth}
        \includegraphics[width=\textwidth]{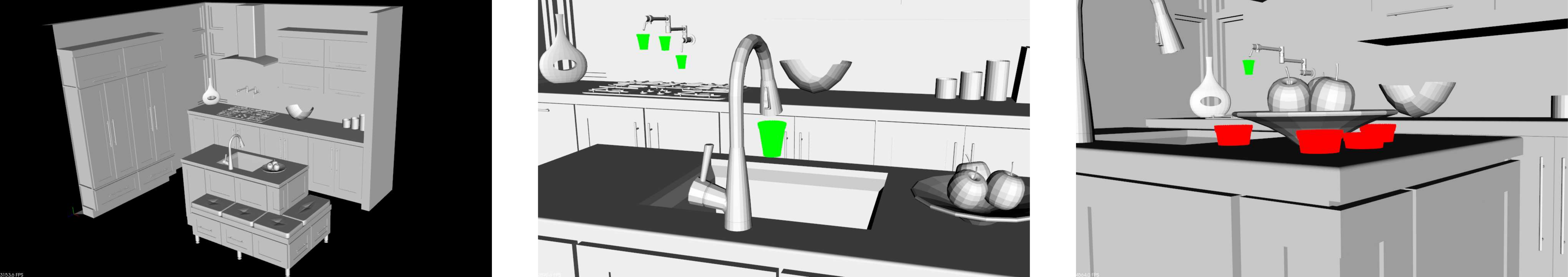}
        \caption{\textit{filling}}
        \label{fig: top naive filling}
    \end{subfigure}
    
    \begin{subfigure}[b]{0.46\textwidth}
        \includegraphics[width=\textwidth]{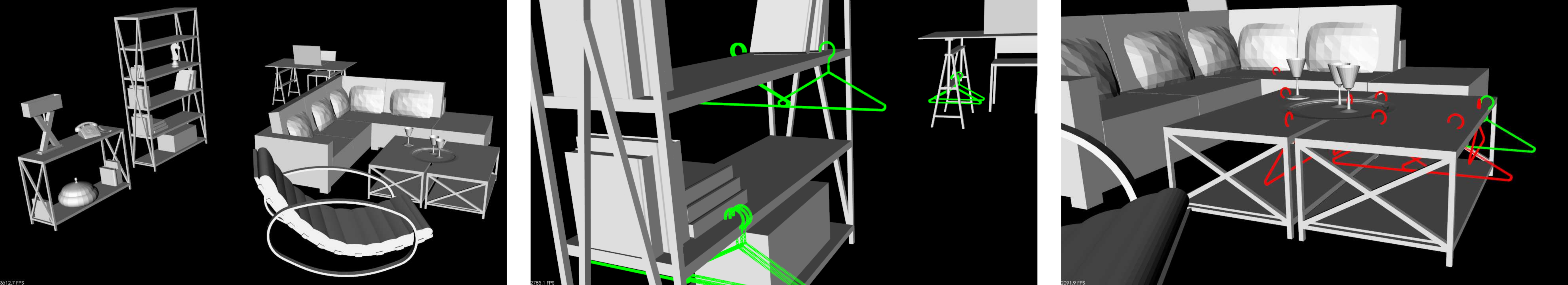}
        \caption{\textit{hanging}}
        \label{fig: top naive hanging}
    \end{subfigure}
    
    \begin{subfigure}[b]{0.46\textwidth}
        \includegraphics[width=\textwidth]{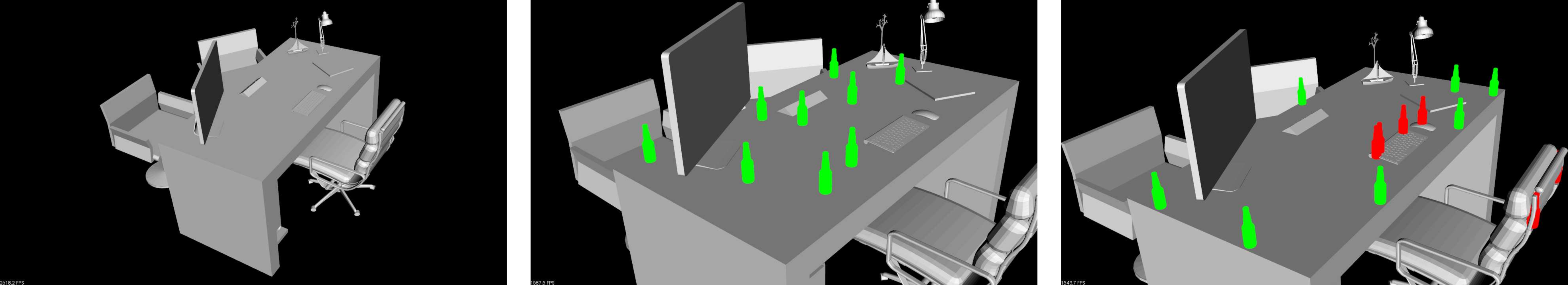}
        \caption{\textit{placing}}
        \label{fig: top naive placing}
    \end{subfigure}
    
    \begin{subfigure}[b]{0.46\textwidth}
        \includegraphics[width=\textwidth]{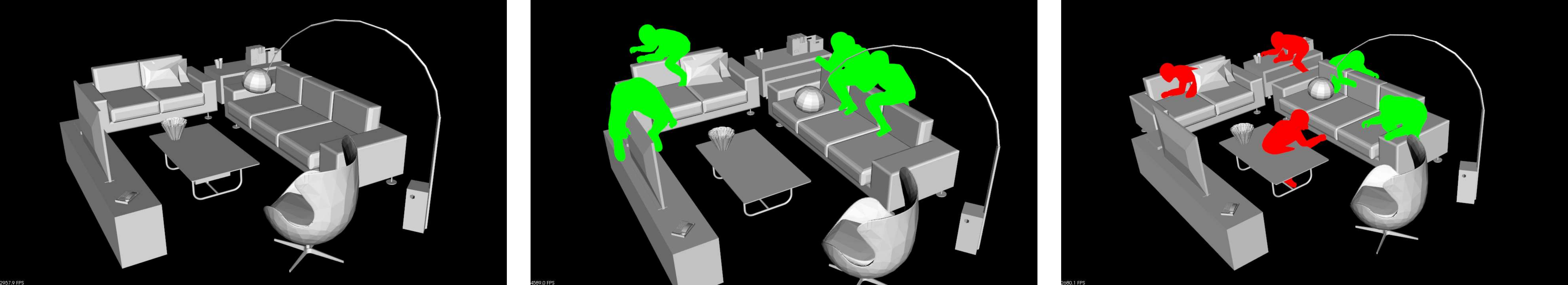}
        \caption{\textit{riding}}
        \label{fig: top naive riding}
    \end{subfigure}
    
    \begin{subfigure}[b]{0.46\textwidth}
        \includegraphics[width=\textwidth]{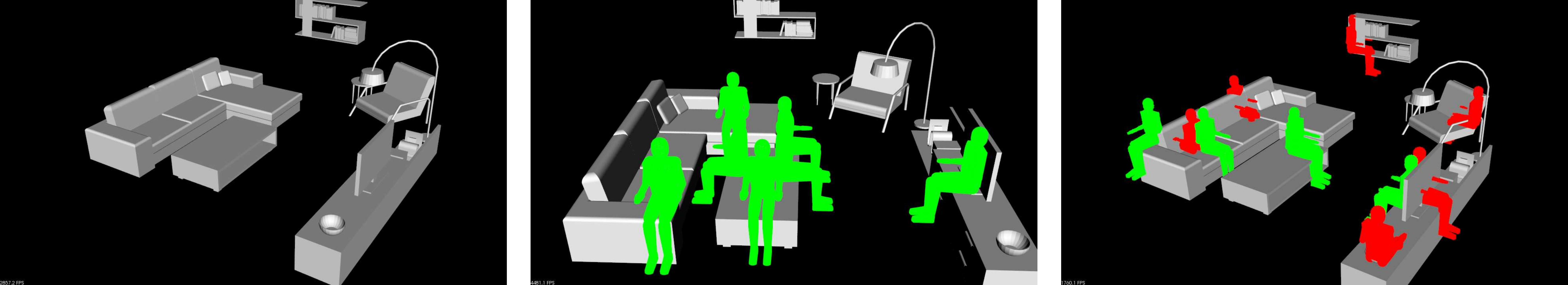}
        \caption{\textit{sitting}}
        \label{fig: top naive sitting}
        \vspace{-2mm}
    \end{subfigure}
    \caption{Affordance predictions. Results on the center column show predicted positions using the iT descriptor. Results in the column on the right show predictions made with the baseline \textit{Naive} algorithm. \textit{Naive} algorithm predicts good locations with equal probability as bad or unachievable configurations (red).}
 	\label{fig: top naive}
    \vspace{-3mm}
\end{figure}

We then evaluated and compared our results against a baseline algorithm that we call \textit{Naive}. This algorithm simply computes a score using ICP between the query object and the scene object, but without any explicit representation of the interaction between the objects; therefore the goal is to find the best possible alignment at test time using the score of the alignment in the interaction example as matching criteria. This is somewhat representative of methods that use object instances as examples instead of instances representing the interaction between objects.
Fig. \ref{fig: top naive} shows results contrasting the \textit{Naive} algorithm and our approach. For fairness, both of the baseline algorithms, sample points and use normal vectors on the scene similarly as we do in our approach.


One of the first things to notice is that the \textit{Naive} approach selects good locations using mainly the normals comparison. While it does find some expected locations, it also predicts as acceptable the locations with object penetrations, occlusions or intersections; these kind of predictions would not be useful or achievable in reality. For instance \ref{fig: top naive sitting} and \ref{fig: top naive riding} show \textit{Naive} predictions for \textit{sitting} and \textit{riding} where the legs or parts of the body (query-object) are inside furniture. Similar cases are observed in Fig. \ref{fig: top naive filling} - \ref{fig: top naive placing}, where the predicted locations would make the query-object collide or to be inside other objects in the scene.

To further evaluate the affordance prediction results, Amazon Mechanical Turk was employed to investigate how acceptable our results are according to human criteria. Human markers were asked to select good locations for each one of the 5 affordances considered in our research. They had to choose amongst different location options, these options consisted of our top 5 and worst 5 predictions per scene, in the expectation that humans would select our top predictions over the bottom ones. A total of 85 human markers participated in the evaluation, each one provided 5 answers (one per affordance). Using this annotation as ground truth, we compute performance metrics for our top 425 predictions. Results of this evaluation are shown in Table \ref{tab: amazon}, which shows that on average our approach achieves a precision of 84.92\% and f-score of 91.17\%; significantly outperforming the baseline methods in nearly all the affordance predictions. In other words, using a single example, our method consistently predicts top geometric affordance locations in unseen areas that agree with human criteria approximately 85 percent of the time; outperforming the baselines by 20-40\% on average.


\begin{figure}[ht]
        \centering
	\includegraphics[width=0.47\textwidth]{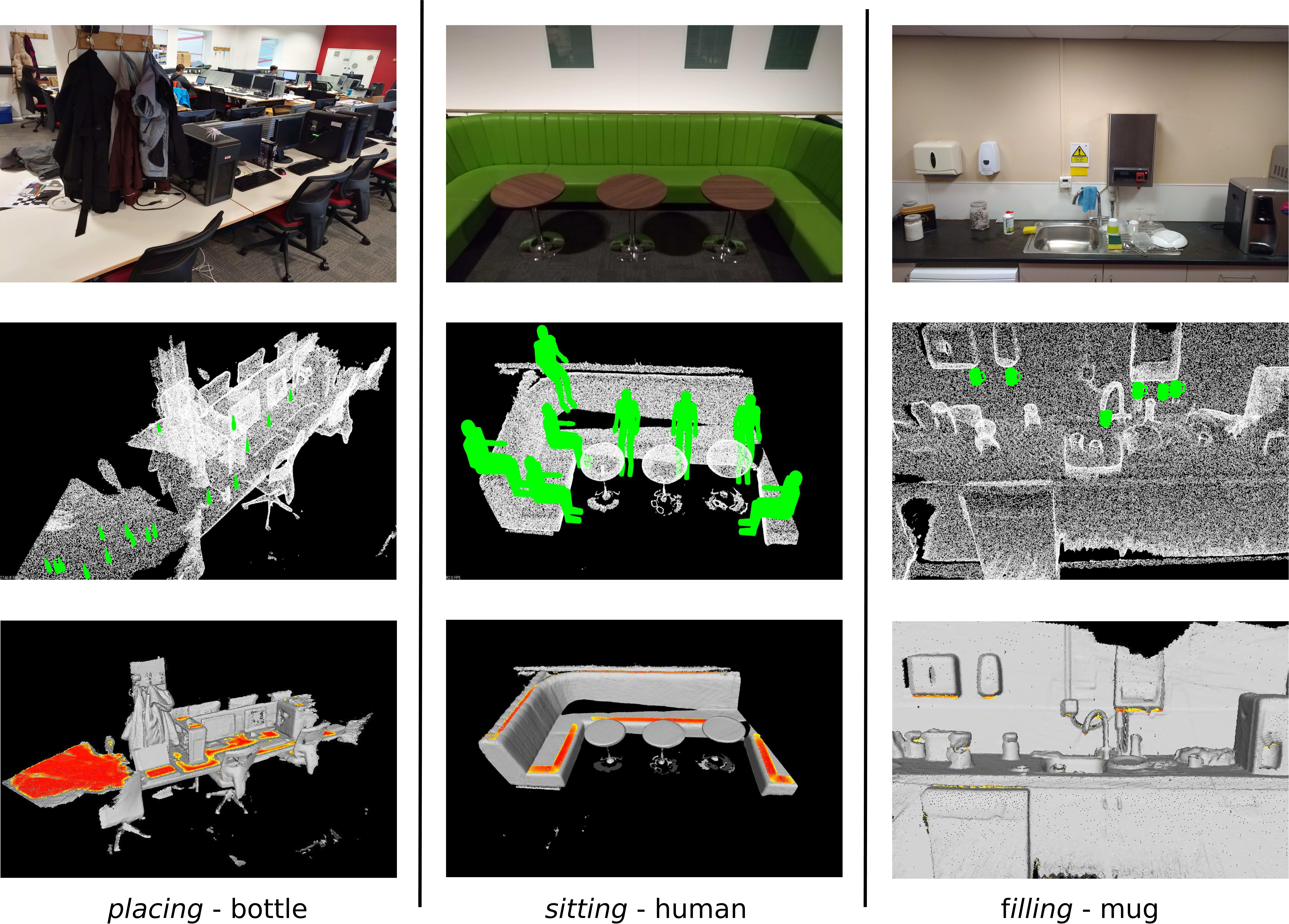}
      \caption{Affordance heatmap with predicted locations in RGBD scenes. From left to right: \textit{placing} a bottle in office environment, \textit{sitting} in reading room and \textit{filling} mug in kitchen. Examples of \textit{riding} motorbike and \textit{hanging} coat hanger in office desk can be seen in Fig. \ref{fig: general view} and Fig. \ref{fig: real hanging} respectively.}
        \label{fig: real-scenes}
        \vspace{-3mm}
\end{figure}

\begin{table}[ht]
\centering
\resizebox{0.47\textwidth}{!}{%
\begin{tabular}{l|c|c|c|c|c|c|}
\cline{2-7}
 & \multicolumn{2}{c|}{iT} & \multicolumn{2}{c|}{Naive} & \multicolumn{2}{c|}{BS} \\ \cline{2-7} 
\multicolumn{1}{c|}{} & Accuracy & F-score & Accuracy & F-score & Accuracy & F-score \\ \hline
\multicolumn{1}{|l|}{placing} & 96.92 & \textbf{98.44} & 3.08 & 5.97 & 3.08 & 5.97 \\ \hline
\multicolumn{1}{|l|}{sitting} & 64.62 & \textbf{78.50} & 64.62 & \textbf{78.50} & 35.38 & 52.27 \\ \hline
\multicolumn{1}{|l|}{filling} & 100 & 100 & 100 & 100 & 100 & 100 \\ \hline
\multicolumn{1}{|l|}{riding} & 92.31 & \textbf{96.00} & 92.31 & \textbf{96.00} & 7.69 & 14.29 \\ \hline
\multicolumn{1}{|l|}{hanging} & 70.77 & \textbf{82.88} & 29.23 & 45.24 & 70.77 & \textbf{82.88} \\ \hline
\multicolumn{1}{|l|}{\textbf{Average}} & \textbf{84.92} & \textbf{91.17} & 57.85 & 64.14 & 43.38 & 51.08 \\ \hline
\end{tabular}
}
\caption{Affordance prediction performance evaluated according to human markers criteria (in terms percentage).}
\label{tab: amazon}
\vspace{-3mm}
\end{table}

It is worth noticing that given human location annotations for \textit{filling}, all the algorithms predict it as good affordance locations. We believe this is mainly due to the distinctive geometry of faucets and sinks, which are usually found very seldom (one in most kitchen scenes) and this makes easier to correctly detect the \textit{filling} affordance. In complex interactions such as \textit{riding}, the BS algorithm is clearly outperformed by the iT descriptor. As explained previously, this algorithm mainly detects affordance at locations with scene geometries very close to the example; since there is no motorbike-like geometry it struggles to predict such affordance; a similar situations occurs for \textit{sitting} affordances. Another remarkable result is \textit{hanging}; according to: human judgment, iT and BS, hanging a coat hanger on edges of flat surfaces is regarded as possible. Traditional methods based on object appearance would fail to detect these cases.

\subsection{RGB-D data}
We conducted experiments on pointclouds captured with a Asus Xtion sensor using a publicly available dense mapping system \cite{shuda2015}. Additionally, we included 5 publicly available scenes from \cite{Choi2016} which contain scans of a real motorcycles and the indoor scene pointclouds from \cite{Zhou2013}, leading to a testing dataset comprised by 20 real scenes. Using the same pipeline explained before, we query object-affordance pairs for each of these scenes using the \textit{training} example from the synthetic training data. The only pre-processing step carried out to these scenes is the ground plane calibration. Fig. \ref{fig: real-scenes} shows affordance heat-maps for these scenes along example of the predicted locations.

\section{Discussion and Conclusion}
\label{conclusions}
This paper presents and evaluates a new tensor field representation to express the {\it geometric} affordance of one object over another. By expanding the bisector surface representation to a richer tensor field, we are able to estimate affordance locations on previously unseen scenes from a single example. The introduction of weighted tensor leads to affordance keypoints that allow faster decisions per query point and a compact and straight forward way to compute a descriptor. Our evaluation is carried out with both synthetic and real RGBD scenes. The performance of our interaction tensor is significantly better in agreeing with crowdsourced opinions than the results of the baseline methods. Overall, we see this work as an effort to motivate further advancing of approaches in Vision which, such as Active Perception 
\cite{Bajcsy2016}, are more {\it ecological} in nature and consider the needs of the perceiving agent.
{\small
\bibliographystyle{ieee}
\bibliography{references}
}

\end{document}